\pdfoutput=1
\documentclass[11pt]{article}

\usepackage[]{ACL2023}

\usepackage{times}
\usepackage{latexsym}

\usepackage{amsmath}
\usepackage{amssymb}
\usepackage{booktabs}
\usepackage{algorithm}
\usepackage{algpseudocode}
\usepackage{graphicx}
\usepackage{adjustbox}
\usepackage{multirow}
\usepackage{subfig}
\usepackage{setspace}

\usepackage[T1]{fontenc}
\usepackage[utf8]{inputenc}

\usepackage{microtype}

\usepackage{inconsolata}

\usepackage{xcolor}
\usepackage{longtable}
\newcounter{myenumi}
\newcounter{myenumii}[myenumi]
\usepackage{etoolbox}

\makeatletter
    \patchcmd{\HyField@FlagsRadioButton}{\HyField@SetFlag{Ff}{Radio}}{}{}{}
\makeatother

\title{Cross-modal Attention Congruence Regularization for\\Vision-Language Relation Alignment}

\author{
    Rohan Pandey\quad
    Rulin Shao\quad
    Paul Pu Liang \\
    \quad
    \textbf{Ruslan Salakhutdinov}\quad
    \textbf{Louis-Philippe Morency}\\
    Language Technologies Institute and Machine Learning Department \\
    Carnegie Mellon University \\
    \texttt{\{rspandey,rulins\}@cs.cmu.edu} \\
}

\begin{document}
\maketitle
\begin{abstract}
Despite recent progress towards scaling up multimodal vision-language models, these models are still known to struggle on compositional generalization benchmarks such as Winoground. We find that a critical component lacking from current vision-language models is relation-level alignment: the ability to match directional semantic relations in text (e.g., `mug \textbf{in} grass') with spatial relationships in the image (e.g., the position of the mug \textbf{relative} to the grass). To tackle this problem, we show that relation alignment can be enforced by encouraging the language attention from `mug' to `grass' (capturing the semantic relation `in') to match the visual attention from the mug to the grass (capturing the corresponding physical relation). Tokens and their corresponding objects are softly identified using a weighted mean of cross-modal attention. We prove that this notion of soft cross-modal equivalence is equivalent to enforcing congruence between vision and language attention matrices under a `change of basis' provided by the cross-modal attention matrix. Intuitively, our approach projects visual attention into the language attention space to calculate its divergence from the actual language attention, and vice versa. We apply our Cross-modal Attention Congruence Regularization (CACR) loss to fine-tune UNITER and improve its Winoground Group score by 5.75 points.
\end{abstract}

\section{Introduction}
\begin{figure}[ht]
    \centering
    \includegraphics[width=\columnwidth]{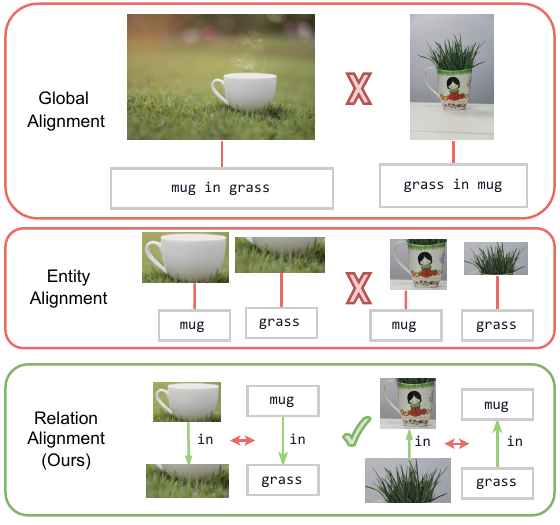}
    \caption{Global Alignment (GA) only aligns the entire image with the corresponding caption. Entity Alignment (EA) extracts entities from the image and caption for finer-grained alignment. Relation Alignment (RA) cross-modally aligns the intra-modal relations between entities in both the image and the text. We show RA is vital to improve compositional performance.}
    \label{fig:wino_overview}
\end{figure}
% Vision-Language compositionality, structural alignment, Winoground task... @Rulin

% - we need RQ2 to explore why attention is specifically interesting in structural training since
    % - there may be other mechanisms for structural training
    % - cacr should generalize to flickr30k (other datasets)
    % - cacr should generalize to other models (vinvl etc)

Compositionality is the ability to combine meanings of constituents according to structured rules. Recent work shows that Vision-Language Models (VLMs) fail to construct compositional representations and generally ignore syntactic \& structural information \cite{thrush22, milewski22, liang2022foundations}.
Winoground \cite{thrush22} is a vision-language compositionality task that tests a VLM's ability to match syntactic permutations of text with their visual interpretations, for example correctly matching ``grass in mug'' and ``mug in grass'' to their corresponding images. Winoground finds that all recent state-of-the-art VLMs perform below chance levels on this compositionality task. Contemporaneously, \citet{milewski22} probe for structural knowledge in VLMs, finding that they encode significantly less linguistic syntax than Language Models (LMs) and virtually no visual structure. Recently, \citet{yuksekgonul2022bags} built a large dataset confirming that VLMs treat images as a `bag of objects' and don't adequately represent visuo-linguistic relations.

Since models must determine whether the compositional structure of an image matches that of the caption, it's important for the model to learn to cross-modally align intra-modal relations. That is, if the relation from `mug' to `grass' is `in-ness', the model should recognize when the equivalent physical relation holds between a mug and grass in the image, and representationally align these relations such that an image-text matching head may more easily determine whether the relations are cross-modally equivalent. In simpler terms, the compositional structure of input for each modality should be represented such that they can be cross-modally matched even for difficult examples like Winoground.

% However, standard contrastive learning and image-text matching training setups don't explicitly seek to cross-modally align compositional structure \cite{}. Instead, they align at a global sentence-image level, which is prone to spurious correlations \cite{yuksekgonul2022bags, yang21catt} and has empirically been shown to result in models forgetting their unimodal structural knowledge \cite{milewski22}. Models trained in this \textbf{global alignment} fashion such as CLIP \cite{radford2021clip} are some of the worst performers on Winoground.

% Much work takes alignment a step deeper, focusing on \textbf{entity alignment} of visual objects and words \cite{li2020oscar, kervadec19weak, guo19vsua, liu2021kdvlp}. These approaches generally perform better, with VinVL \cite{zhang2021vinvl} being the prior state-of-the-art. Unfortunately, even entity alignment is not enough to solve Winoground because example pairs in this task both contain the same words \& objects—it is the relations between the entities that differ.

Unfortunately, there has been less highly influential work on \textbf{relation alignment} between vision \& language, and \citet{thrush22} did not benchmark any such models. In this work, we begin exploration of these relation alignment approaches by tentatively grouping them into 3 categories: 
\begin{enumerate}
    \item Structural Data: training a model on data that explicitly captures relational structure
    \item Structural Model: infusing an inductive bias into the architecture of the model that enables more compositional representations
    \item Structural Training: modifying the objective function or imposing a parameter constraint to encourage relation alignment
\end{enumerate}

Since Structural Data approaches require complex annotations and Structural Model approaches are often incompatible with large transformers, we identify Structural Training as a promising avenue for providing compositional inductive biases to VLMs due to their architecture-agnostic compatibility and computational scalability.

In this work, we propose a Structural Training approach for relation alignment that uses the cross-modal attention matrix as a change of basis\footnote{not defined in a strict linear algebraic sense} to the opposite modality, which we then compare to the original modality to calculate a divergence loss, effectively measuring cross-modal congruence between intra-modal attentions. 

We show how our approach, Cross-modal Attention Congruence Regularization (CACR), generalizes previous Structural Training work on cross-modal attention regularization (IAIS \cite{ren21iais}) by taking into account all possible entity alignments and computationally simplifying relation alignment. The CACR regularization term can easily be dropped into most transformer-based Vision-Language model objectives with no added data and minimal computational overhead, to encourage relation alignment during training. Finally, we show that CACR$_{\text{base}}$ improves on IAIS$_{\text{base}}$—where IAIS$_{\text{large}}$ holds the current state-of-the-art on Winoground.

% This is in effect a measurement of congruence$^1$ between each modality's intra-modal attentions and the opposite modality's corresponding attention values.
% Although it isn't strictly a regularizer since it is calculated from attention values for each example rather than the underlying parameters, we refer to it as a regularizer rather than auxiliary loss in the vein of prior Structural Training work \cite{ren21iais} and to provide a better intuition of its self-supervised nature.

% At a higher level, attention values in transformers may be seen as an informational gating mechanism that implicitly encode how representations are composed. For example, past work in language has shown how syntax trees may be extracted \cite{marevcek2019balustrades, pandey22synnamon} from attention across layers and used to guide attention \cite{bai2021syntax, li2020improving} for improved compositionality. Our work extends this intuition to the multimodal domain by showing that we can use the cross-modal attentions, which as a change of basis matrix encode a transformation from one modality's compositional structure to the opposite modality's, to encourage relation alignment between the modalities. 

\section{Related Work}

Below, we categorize several relation alignment approaches following the framework in Sec. 1. 
\begin{enumerate}
    \item Structural Data \cite{Wu_2019_univse, zhang2019agha, yu20ernievil, cui21rosita, wan2021cliora, khan2022simla}
    \item Structural Model \cite{andreas2016neural, guo19vsua, hong2021vlgrammar, zhang-2022-improve, wang2022sgeitl, kim2022cross, wang22vqagnn}
    \item Structural Training \cite{ren21iais, yang21apn, yang21catt, xue2021imf}
\end{enumerate}

While some of these works introduce ideas from multiple of these categories, we group them by their core contribution. For example, ROSITA proposes a graphical data pre-training approach, and a self-supervised objective to accompany it; we consider it a Structural Data approach since the training objective ultimately is just a necessity for the data being provided.

Unfortunately, many of these works do not provide publicly available code or pre-trained checkpoints, so we were unable to complete an exhaustive analysis of the compositional performance of these relation alignment approaches. Due to the added complexity of Structural Model approaches, we leave exploration of their compositional abilities to future work.

Regardless, we chose one exemplar for both Structural Data (ROSITA) and Structural Training (IAIS) that made their pre-trained image-text matching checkpoints available; we generated their scores on Winoground, which have not previously been calculated. In Tab. \ref{table:wg_alignment}, we present these two relation alignment models' Winoground scores alongside a few entity alignment and global alignment models.

\begin{table}[h]
\centering
\begin{tabular}{ p{2.8cm}||p{1cm} p{1cm} p{1cm} }
 \hline
 Model &Text &Image &Group\\
 \hline
 MTurk Human & 89.50 & 88.50 & 85.50\\
 \hline
 IAIS (RA-ST) & $\mathbf{42.50}$ & $\mathbf{19.75}$ & $\mathbf{16.00}$\\
 OSCAR+ (EA) & 37.75 & 17.75 & 14.50\\
 ROSITA (RA-SD) & 35.25 & 15.25 & 12.25\\
 UNITER (EA) & 38.00 & 14.00 & 10.50\\
 CLIP (GA) & 30.75 & 10.50 & 8.00\\
 LXMERT (GA) & 19.25 & 7.00 & 4.00\\
 \hline
\end{tabular}
\caption{Comparison of Winoground scores for models \cite{ren21iais, zhang2021vinvl, cui21rosita, chen2020uniter, radford2021clip, tan2019lxmert} using Global Alignment (GA), Entity Alignment (EA), Relation Alignment with Structural Data (RA-SD), and Relation Alignment with Structural Training (RA-ST). We find that IAIS, a recent relation alignment approach that uses attention regularization for structural training achieves universal performance improvements.}
\label{table:wg_alignment}
\end{table}

Notice that global alignment approaches tend to perform the lowest on Winoground, even when scaled considerably. Entity alignment approaches perform intermediately and OSCAR+ specifically held the state-of-the-art prior to our benchmarking of these relation alignment models. Of the two relation alignment approaches we benchmark, IAIS beats out OSCAR+ and achieves a new state-of-the-art on Winoground. But ROSITA, despite providing structural data to encourage cross-modal relation alignment, underperforms OSCAR+. We attribute this partly to the improved visual features OSCAR+ has access to as a result of VinVL, but further comparison of IAIS and ROSITA is explored in our recent work.

Based on these past results and analysis, we choose to further explore structural training approaches to relation alignment. In other words, our research question becomes: How can we infuse the vision-language model's training objective with an implicit structural prior that encourages cross-modal alignment of relations?

\section{Cross-modal Attention Congruence Regularization}
% \rulin{We may re-organize this section into three parts (each with a paragraph title):
% 1. explain why we could use attention alignment to achieve relation alignment.
% 2. propose the soft cross-modal equivalence and explain its relation with hard cross-modal equivalence.
% 3. write the CACR equation, say it's a computationally efficient form of soft cross-modal equivalence, and can be easier parallelized, more efficient, and better aligned than hard alignment.
% }

\begin{figure*}
    \centering
    \includegraphics[width=\linewidth]{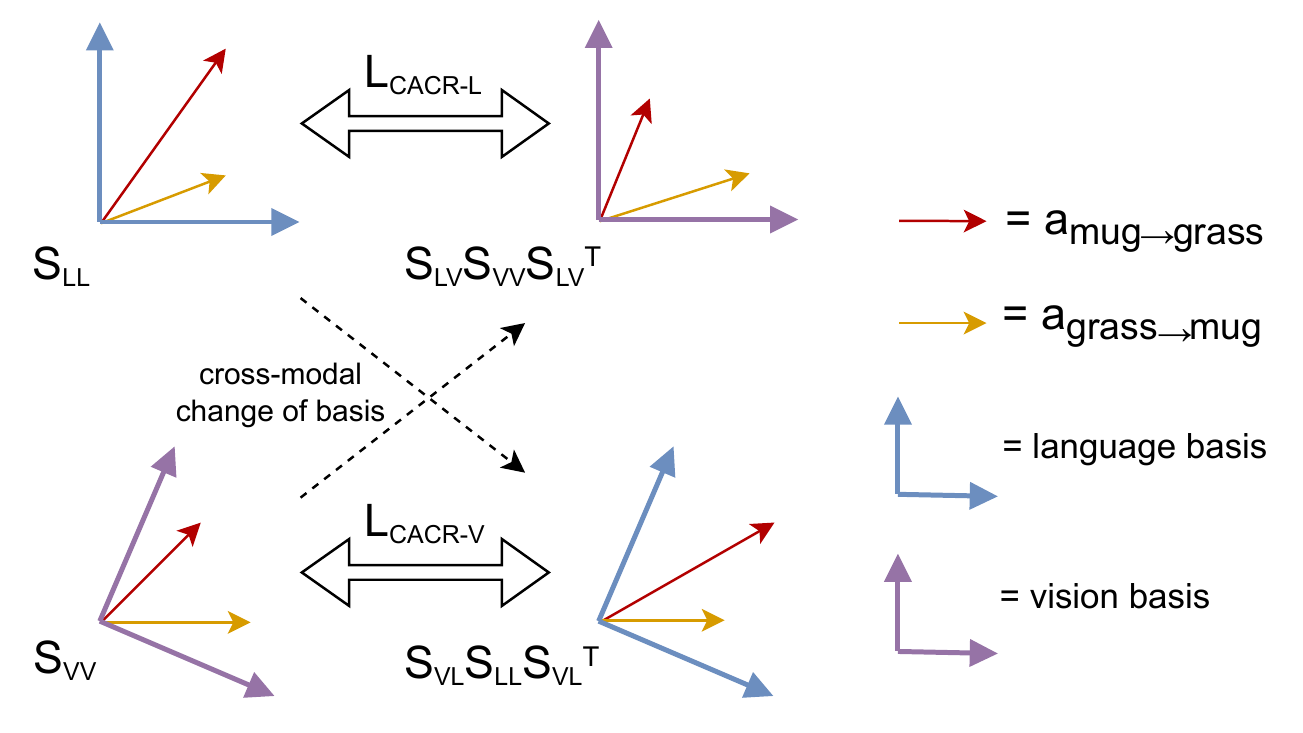}
    \caption{Top: language attention ($S_{LL}$) is aligned with the visual attention projected into the language basis ($S_{LV}S_{VV}S_{LV}^\top$) to calculate $\mathcal{L}_{CACR-L}$; specific attention values (yellow, red) capturing intra-modal relations are cross-modally aligned as a result. Bottom: as above, but in the vision basis.}
    \label{fig:basis}
\end{figure*}

% TODO: give roadmap of Rel Work section

To attempt a solution to this question, we begin by noting that attention activations encode some degree of relational information. Attention values in transformers may be seen as an informational gating mechanism that implicitly encode how representations are composed \cite{abnar2020quantifying}. For example, past work in language has shown how syntax trees may be extracted \cite{marevcek2019balustrades} from attention across layers and used to guide attention \cite{bai2021syntax, li2020improving} for improved compositionality. In this section, we extend this intuition to the multimodal domain by proposing to use the cross-modal attentions, which as a change-of-basis matrix encode a transformation from one modality's compositional structure to the opposite modality's, to encourage cross-modal relation alignment.

\subsection{Relation Alignment Using Attention}
In specific, we focus on the self-attention matrix $S$ computed in a transformer by 
\begin{equation}
    S = QK^\top = (X W^Q)(X W^K)^\top
\end{equation}

Then, some row $i$ in S corresponds to a distribution over columns $j_0, ..., j_n$ where $S_{i,j}$ tells us how much of the previous layer's entity representation $j$ we want to infuse into the current layer's entity representation $i$, intuitively their compositional relation. Since $X$ is a series of visual and linguistic tokens, we can segment $S$ into four submatrices for intra- and cross-modal relations \cite{bugliarello-etal-2021-multimodal}. Denote the intra-modal attention submatrices in the last multimodal encoder layer as $S_{VV}$ (vision to vision) and $S_{LL}$ (language to language); the cross-modal attention matrices as $S_{VL}$ (vision to language) and $S_{LV}$ (language to vision).

\begin{equation}
    S =
\begin{pmatrix}
S_{LL} & S_{LV} \\
S_{VL} & S_{VV}
\end{pmatrix}
\end{equation}

If an image and caption have the same underlying compositional structure, the entities that cross-modally correspond to each other should bear similar intra-modal compositional structure. That is, a word $w$ should attend to other words (in $S_{LL}$) in a similar way that its visual object counterpart $o$ attends to other objects (in $S_{VV}$). Furthermore, we can use the cross-modal matrices ($S_{LV}$ and $S_{VL}$) to identify entities that cross-modally correspond as they will generally attend to each other \cite{aflalo2022interpret}. Unfortunately, since representations are heavily contextualized by the final layer, clear bijective correspondences between words and objects may not always be identified using an argmax over the cross-modal attention matrix as \citet{ren21iais} attempts. Deeper analysis of when their model, IAIS, fails to identify cross-modal bijective correspondences is provided in Sec. \ref{sec:analysis}.

\subsection{Attention Congruence} 

We opt to use the cross-modal matrices ($S_{LV}$ and $S_{VL}$) as a whole to `change basis' to the opposite modality, with which we can then calculate `congruence' with the original modality. However, we use `change of basis' and `congruence' loosely since the cross-modal matrices are not guaranteed to be square and thus do not satisfy strict linear algebraic definitions. We formulate $S_{VV}$ in the language basis as $S_{LV}S_{VV}S_{LV}^\top$, which we then encourage to be similar to $S_{LL}$.

% TODO: clean up paragraph below. not sure we need to discuss this under the hood stuff yet. maybe instead provide lin alg intuition on congruence. or talk abt how this projects *relations* cross-modally.
Under the hood, this says that for each $a_{i \rightarrow j} \in S_{LL}$, we can use row vectors $S_{LV,i}$ and $S_{LV,j}^\top$ to calculate a weighted sum $a_{i \rightarrow j}^*$ over $S_{VV}$. If we were to do this for all $i,j$, we would construct a matrix of the same shape as $S_{LL}$ where each entry is $a_{i \rightarrow j}^*$, i.e. an approximation of the visual correspondent of the relation $a_{i \rightarrow j}$ taking into account all the possible cross-modal alignments of $i$ and $j$. Since this computation intuitively makes a lot of sense and may more easily be compared to previous approaches, we choose to illustrate it in Fig. \ref{fig:attention}. However, since this computation is relatively expensive, we instead use the $S_{LV}S_{VV}S_{LV}^\top$ formulation which produces the same matrix of $a_{i \rightarrow j}^*$ values but with considerably fewer operations. This also enables us to view the operation as a `change-of-basis' to the opposite modality and the CACR loss as encouraging a sense of cross-modal `congruence'.

Specifically, we align the original $S_{LL}$ with the language-basis $S_{VV}$ matrix using $\mathcal{L}_{\text{CACR-L}}$:
\begin{equation}
    \mathcal{L}_{\text{CACR-L}} = \text{m-KL}(\sigma(S_{LV}S_{VV}S_{LV}^\top), \sigma(S_{LL})).
\end{equation}

We apply a softmax to normalize both matrices since $S_{LV}S_{VV}S_{LV}^\top$ will generally be larger in scale due to summation. Additionally, m-KL$(\cdot)$~\cite{ren21iais} is a symmetric matrix-based Kullback-Leibler Divergence (m-KL) which measures the distance between two matrices $S$ and $S'$:
\begin{equation}
    \text{m-KL}(S, S') = \sum_i^N \text{KL}(S_i||S_i') + \text{KL}(S_i'||S_i),
\end{equation}
where $(\cdot)_i$ stands for the $i^{\text{th}}$ row-vector in the matrix.

Similarly, we have $\mathcal{L}_{\text{CACR-V}}$:
\begin{equation}
    \mathcal{L}_{\text{CACR-V}} = \text{m-KL}(\sigma(S_{VL}S_{LL}S_{VL}^\top), \sigma(S_{VV})),
\end{equation}

Combining $\mathcal{L}_{\text{CACR-V}}$ and $\mathcal{L}_{\text{CACR-L}}$, we present our $\mathcal{L}_{\text{CACR}}$ objective, an attention activation regularizer for cross-modal relation alignment:
\begin{equation}\label{eq:loss}
    \mathcal{L}_{\text{CACR}} = \mathcal{L}_{\text{CACR-V}} + \mathcal{L}_{\text{CACR-L}}.
\end{equation}

When the vision inputs and the language inputs have the same sequence length and $S_{VL}, S_{LV}$ are invertible, then $S_{VV}$ and $S_{VL}S_{LL}S_{VL}^\top$ (as well as $S_{LL}$ and $S_{LV}S_{VV}S_{LV}^\top$) can become strictly congruent. In this case, $S_{VL}S_{LL}S_{VL}^\top$ can be interpreted as the language view of $S_{VV}$. Aligning $S_{VL}S_{LL}S_{VL}^\top$ and $S_{VV}$ leads to cross-modal relation alignment. It is similar for $S_{LV}S_{VV}S_{LV}^\top$ and $S_{LL}$. In the general case where the vision inputs and the language inputs may have different sequence lengths, the two forms are not linear algebraically congruent but the relevant intuition still holds.

\subsection{Hard and Soft Cross-modal Equivalence}

\begin{figure*}
    \centering
    \includegraphics[width=\linewidth]{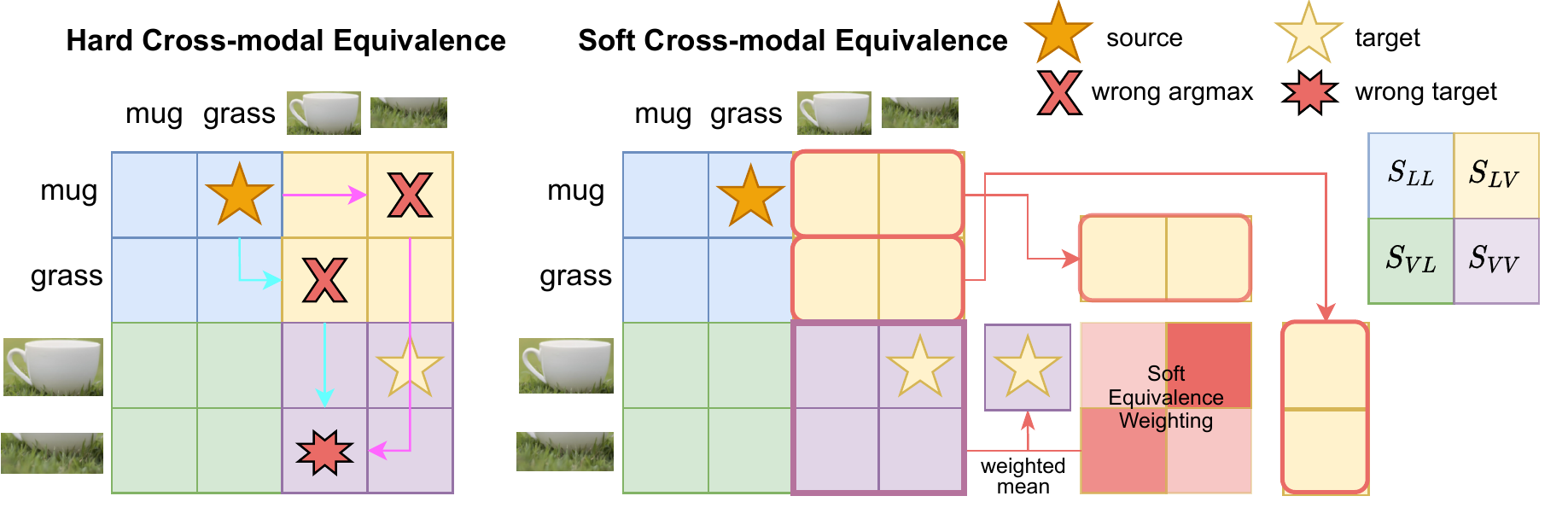}
    \caption{Comparison of the hard cross-modal equivalence used in IAIS (left) and the soft cross-modal equivalence we propose in CACR (right), with an example from Winoground to illustrate how the target visual equivalent (yellow star) of a source linguistic relation (orange star) is calculated. Hard cross-modal equivalence attempts to build a one-to-one mapping between language and vision by applying argmax (red crosses) to the $S_{LV}$ row vectors. Our soft cross-modal equivalence instead uses the whole $S_{LV}$ row vectors to calculate a weighted (red) mean over $S_{VV}$. The scalar that is produced corresponds to the attention from `mug' to `grass' but in a visual basis. We note that IAIS can be seen as a special case of soft cross-modal equivalence by forcing the attention matrix (red) to be a one-hot matrix where the max value is set to 1 and all others to 0. CACR linear algebraically simplifies soft cross-modal equivalence for computational efficiency.}
    \label{fig:attention}
\end{figure*}

In this section, we show that CACR can be interpreted as leveraging cross-modal soft equivalences, where IAIS \cite{ren21iais} uses hard bijective equivalences. In their approach, each element in the intra-modal attention matrix is aligned with a single counterpart in the opposite modality. This is built upon a strict assumption that there exists a one-to-one mapping (provided by an argmax over the cross-modal attention) from $S_{LL}$ to $S_{VV}$ and vice versa, which is unsatisfied in practical cases. CACR may be seen as a soft cross-modal equivalence method which instead uses the whole $S_{LV}$ (or $S_{VL}$) to implicitly build an `equivalence weighting' which is then used to compute a weighted mean over $S_{VV}$ (or $S_{LL}$). We illustrate and compare hard cross-modal equivalence and our soft cross-modal equivalence in Figure~\ref{fig:attention}, taking the language-side alignment as an example. 

We note that IAIS could be seen as a special case of soft cross-modal equivalence by forcing the cross-modal attention map to be a one-hot matrix, i.e., taking the argmax of the attention matrix as the index of the cross-modal counterpart. We show in Section~\ref{sec:analysis} that IAIS can have inferior performance when a clear bijective cross-modal correspondence isn't available.

In Alg.~\ref{alg:siaisl}, we show the pseudo-code of the soft cross-modal equivalence method for calculating the vision-side loss. $\mathcal{S}_{CACR-V}$ can be computed similarly. Computing the hard and soft cross-modal equivalence is computationally complex and difficult to be parallelized due to indexing operations. For practical applications, we sought to simplify this soft cross-modal equivalence algorithm to a mathematical equivalent that would improve computational tractability. From here, we arrive at CACR, which is a closed-form formulation of soft cross-modal equivalence which utilizes only differentiable matrix multiplications. Therefore, our CACR is more \textbf{computationally efficient} and \textbf{easier to be parallelized} than soft cross-modal equivalence. 

\begin{algorithm}
\setstretch{1.2}
\caption{Soft Cross-modal Equivalence (V)}
\label{alg:siaisl}

\begin{algorithmic}[1]
\Require $S_{LL} \in N \times N, S_{VL} \in N \times M, S_{VV} \in M \times M$
\State $\mathcal{L} \gets 0$
\For {$i, j \in S_{VV}$}
\State $W \gets S_{VL}[i] \cdot S_{VL}^\top[j]$ \Comment{soft weighting}
\State $a_{i \rightarrow j}^* \gets \overline{W \circ S_{LL}}$ \Comment{element-wise weighted mean}
% \State $E \gets (A - S_{VV})^2$x
% \State $L \gets \Sigma(E \cdot P_{i,j})$
\State $\mathcal{L} = \mathcal{L} + \text{m-KL}(a_{i \rightarrow j}^*, S_{VV}[i,j])$
\EndFor
\State \Return $\mathcal{L}$
\end{algorithmic}
\end{algorithm}

\subsection{Proof of Equivalence Between CACR and Soft Cross-modal Equivalence}
Computing the hard (IAIS) and soft cross-modal equivalence is computationally complex and difficult to parallelize due to indexing operations. However, CACR loss is mathematically equivalent to soft cross-modal equivalence but can be computed efficiently. We take $\text{CACR}_{V}$ for illustration of this equivalence, but $\text{CACR}_{L}$ can be proved in the same way.

Beginning with the visual-basis form of $S_{LL}$ in CACR, the attention at index $[i, j]$ in $S_{VL}S_{LL}S_{VL}^\top$ is
\begin{equation}\label{eq:proof}
    \begin{array}{l}
        (S_{VL}S_{LL}S_{VL}^\top)[i,j] \\[4pt]
        
        = \sum_p^{N_L} \sum_k^{N_L} a_{v_i\rightarrow l_k} a_{l_p \rightarrow l_k} a_{v_j \rightarrow l_p}  \\[4pt]
        
        = \sum_p^{N_L} \sum_k^{N_L} \underbrace{S_{VL}[i, k] S_{VL}[j, p]}_{\text{soft weighting}} S_{LL}[p, k]
    \end{array}
\end{equation}

% \begin{equation}\label{eq:proof}
%     \begin{array}{l}
%         \sum_p^{N_V} \sum_k^{N_V} \underbrace{S_{LV}[i, k] S_{LV}[j, p]}_{\text{soft weighting}} S_{VV}[p, k] \\[4pt]
        
%         = \sum_p^{N_V} \sum_k^{N_V} a_{l_i\rightarrow v_k} a_{v_p \rightarrow v_k} a_{l_j \rightarrow v_p}  \\[4pt]
        
%         = (S_{LV}S_{VV}S_{LV}^T)[i,j]
%     \end{array}
% \end{equation}

where $a_{v_i\rightarrow l_j}$ stands for the attention from the $i$-th visual token to the $j$-th linguistic token, $N_L$ is the total number of language tokens and $N_V$ is the total number of the visual tokens. Comparing Eq.\ref{eq:proof} and Alg.\ref{alg:siaisl}, we observe that the summation we arrive at above is equivalent to the content of the for-loop (line 3-5). Thus, although of seemingly different linear algebraic form, CACR generalizes IAIS by way of its equivalence to the Soft Cross-modal Equivalence formulation presented above.

\section{Results}

How does CACR compare to other vision-language models in its compositional ability?

We fine-tuned CACR on Flickr30k \cite{young2014flickr} for 5000 epochs using PyTorch \cite{paszke2019pytorch} with a train-validation-test split of 90-5-5. The training batch size is 4 and 31 negative samples are provided for every individual positive sample in a standard image-text matching training setup. We use a learning rate of $5 \times 10^{-5}$, the $AdamW$ optimizer \cite{loshchilov2017decoupled}, and introduce $\mathcal{L}_{\text{CACR}}$ with an exponential warmup schedule. Training was completed on a node with 4 NVIDIA GTX 1080 Ti's, each with 11 GB of memory.

\begin{table}[h]
\centering
\begin{tabular}{ p{2.5cm}||p{1cm} p{1cm} p{1cm} }
 \hline
 Model &Text &Image &Group\\
 \hline
 MTurk Human & 89.50 & 88.50 & 85.50\\
 % Random Chance & 25.00 & 25.00 & 16.67\\
 \hline
 % CACR$_{\text{large}}$ \\
 % IAIS$_{\text{large}}$ & $42.50$ & $19.75$ & $16.00$\\
 % OSCAR+ & 37.75 & 17.75 & 14.50\\
 CACR$_{\text{base}}$ & 39.25 & \textbf{17.75} & \textbf{14.25}\\
 UNITER$_{\text{large}}$ & \textbf{43.50} & 14.75 & 13.75\\
 IAIS$_{\text{base}}$ & $37.50$ & $16.75$ & 13.00\\
%  ROSITA$_{\text{Flickr30k}}$ & 35.25 & 15.25 & 12.25\\
 UNITER$_{\text{base}}$ & 32.75 & 11.75 & 8.50 \\
%  CLIP (ViT-B/32) & 30.75 & 10.50 & 8.00\\
 \hline
\end{tabular}
\caption{CACR outperforms its pre-trained baseline (UNITER) and an alternative attention regularization approach (IAIS) across all Winoground scores.}
\label{table:wg_scores}
\end{table}

In Tab. \ref{table:wg_scores}, we present our approach's scores alongside a few other models. Since we use CACR to fine-tune UNITER, we include scores for the two baseline UNITER sizes. We also include scores for $\text{IAIS}_{\text{base}}$ which is also built on UNITER. 
% However, we only train $\text{CACR}_{\text{base}}$ since $CACR_{\text{large}}$ would've required 8 GPU's on a single node or a cluster with multi-node training—resources we do not have access to.

The fact that $\text{CACR}_{\text{base}}$ outperforms $\text{IAIS}_{\text{base}}$ suggests that, with adequate computational resources, $\text{CACR}_{\text{large}}$ could similarly outperform $\text{IAIS}_{\text{large}}$, potentially achieving a new state-of-the-art on Winoground. Furthermore, its performance compared to $\text{UNITER}_{\text{large}}$ is impressive considering that $\text{CACR}_{\text{base}}$ is approximately half its size in parameters. Despite our resource constraints, we were able to train a partly frozen (first 12 layers) version of $\text{CACR}_{\text{large}}$, which achieves $\text{IAIS}_{\text{large}}$ levels of performance on Winoground (text: 37.5, image: 18.75, group: 15.75) with just 20\% of the training time.

\begin{table}[h]
\centering
\begin{tabular}{ p{1.7cm}||p{0.9cm} p{0.9cm} p{0.9cm} p{0.9cm}}
 \hline
 Model &Image R@1 &Image R@10 &Text R@1 &Text R@10\\
 \hline
 % CACR$_{\text{large}}$ \\
 % IAIS$_{\text{large}}$ & 76.86 & 95.72 & 88.30 & 99.40 \\
 % UNITER$_{\text{large}}$ & 73.56 & 96.76 & 87.30 & 99.20 \\
 IAIS$_{\text{base}}$ & 73.54 & 96.32 & 86.10 & 99.10 \\
 UNITER$_{\text{base}}$ & 72.52 & 96.08 & 85.90 & 98.80 \\
 CACR$_{\text{base}}$ & 70.88 & 95.68 & 83.50 & 98.80 \\
 \hline
\end{tabular}
\caption{CACR performance on Flickr30k has marginal reductions from UNITER, suggesting performance could be improved even further with a hyperparameter search.}
\label{table:flickr_scores}
\end{table}

Finally, we report Flickr30k retrieval scores in Tab. \ref{table:flickr_scores} to verify that we are not somehow overfitting to Winoground. Though CACR takes some minor losses to its retrieval scores, this may be attributed to imperfect hyperparameters, suggesting that CACR's performance on Winoground could be even higher with adequate hyperparameter tuning. It's also important to remember here that we're only training on Flickr30k, so this isn't a case of our model overfitting to Winoground and `forgetting' its true image-text matching ability. Rather, it shows that the hyperparameters that we adapted from IAIS need to be modified to more perfectly train CACR on Flickr30k, which would then carry over to compositional improvements on Winoground.

\section{Analysis}
\label{sec:analysis}
Why does CACR's soft cross-modal equivalence approach outperform hard cross-modal equivalence?

% \subsection{Structural Training}

% % NOTE: from workshop paper?

% Considerable work has explored how providing structural data like scene graphs or syntax parses to vision-language models could encourage relation alignment by explicitly providing such representations. 

\subsection{Qualitative}

Hard cross-modal equivalence, implemented by IAIS, assumes that cross-modal submatrices can be used to find a singular equivalent of an entity in the opposite modality. Specifically, if $i^* = \text{argmax}(S_{LV}[i])$ then $S_{LL}[i]$ should correspond to $S_{VV}[i^*]$. In simple terms, IAIS says the following: if word A attends most to object A and word B attends most to object B, then word A should attend to word B in a similar way that object A attends to object B. Underlying IAIS is the hard assumption that argmaxing over the cross-modal attention submatrix is an effective means of identifying the opposite modality equivalent of an entity. However, we show in this section that this is often not the case. 

\begin{figure}%
    \centering
    \subfloat[\centering \textit{clouds}]{{\includegraphics[width=0.49\columnwidth]{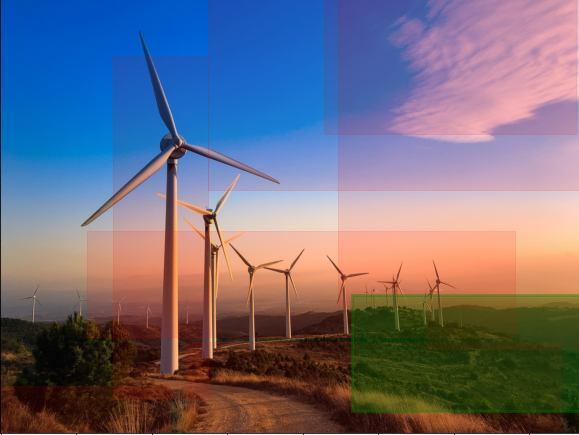} }}%
    \subfloat[\centering \textit{turbines}]{{\includegraphics[width=0.49\columnwidth]{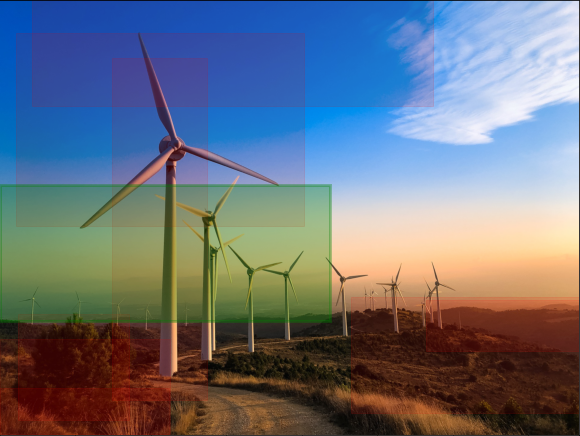} }}%
    
    \subfloat[\centering \textit{``hammering something together"}]{{\includegraphics[width=0.49\columnwidth]{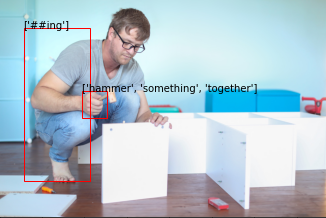} }}
    \subfloat[\centering \textit{``together hammering something"}]{{\includegraphics[width=0.49\columnwidth]{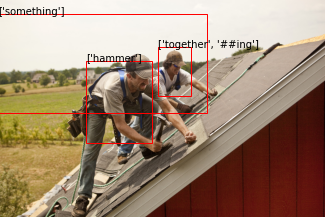} }}

    \caption{Top: UNITER $S_{LV}$ attention for caption \textit{``a few clouds and many wind turbines"}, with the bounding box maximally attended to by the token in green; other highly attended boxes in red. Bottom: UNITER $S_{LV}$ attention with bounding boxes labeled with the tokens that maximally attend to them. Note that argmaxes often fail to precisely identify cross-modal equivalence.}
    \label{fig:slv_ex}
\end{figure}

Given the argmaxes for rows in the $S_{LV}$ submatrix, we can identify the bounding box that each token maximally attends to, which IAIS assumes is its visual equivalent. In Fig. \ref{fig:slv_ex}a, we visualize an example where `clouds' maximally attends (green) to the ground, which would prevent IAIS from identifying the correct cross-modal equivalence. `Turbines' (Fig. \ref{fig:slv_ex}b), on the other hand, maximally attends to a bounding box that better matches our intuition. It is qualitatively clear from the several examples displayed that the argmax assumption often fails to identify the correct cross-modal equivalence. Since words may attend to several visual tokens for different reasons, we shouldn't assume that the cross-modal argmax provides us with a clear bijective correspondence.

% \begin{figure}
%     \centering
%     \includegraphics[width=3.2in]{figs/313353753_2346367778874808_7878248725717376462_n.png}
%     % \includegraphics[width=3.2in]{figs/313347947_1541459683036189_2420796534999527118_n.png}
%     \includegraphics[width=3.2in]{figs/313381154_530533688447659_8940014491868541870_n.png}
%     \caption{IAIS cross-modal argmax entity alignment for \textit{``together hammering something"} and \textit{``a few clouds and many wind turbines"}}
%     \label{fig:alignment}
% \end{figure}

Instead, the cross-modal matrices should be seen as providing useful high-level information about what visual entities are relevant to a word, and vice versa (as intuitively demonstrated by \cite{ilinykh-dobnik-2022-attention}). We can certainly gain useful information about cross-modal correspondences using it, but it isn't as simple as using an argmax, due to words having multiple referents and entity representations being intermixed. Instead, our soft cross-modal equivalence approach takes all the possible cross-modal alignments into account with a weighted sum.

\begin{figure}[ht]
    \centering
    \subfloat[\centering \textit{dog}]{{\includegraphics[width=0.49\columnwidth]{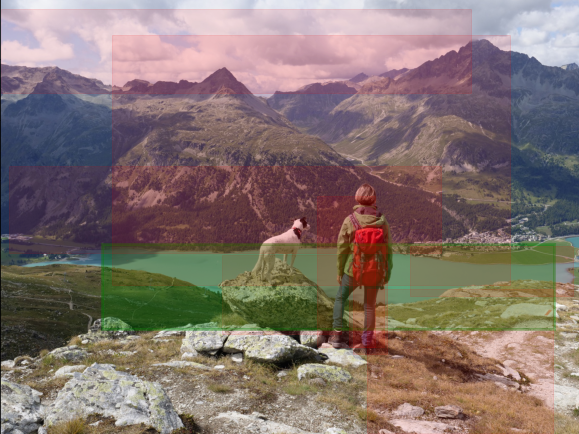} }}%
    \subfloat[\centering \textit{person}]{{\includegraphics[width=0.49\columnwidth]{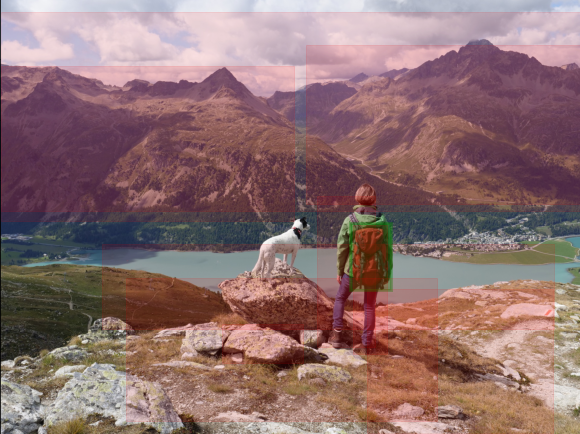} }}%

    \subfloat[\centering  \textit{mug}]{{\includegraphics[width=0.49\columnwidth]{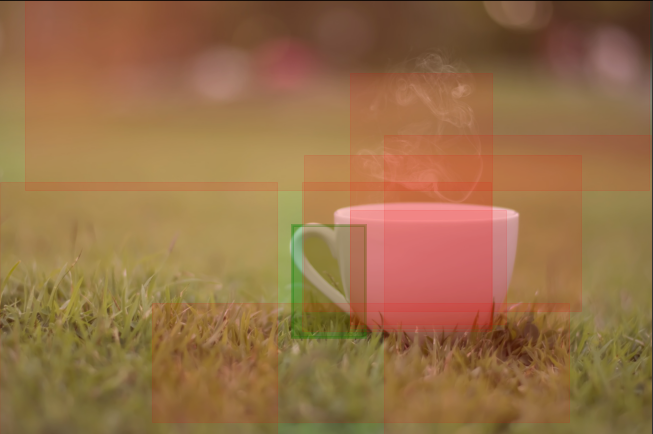} }}
    \subfloat[\centering  \textit{grass}]{{\includegraphics[width=0.49\columnwidth]{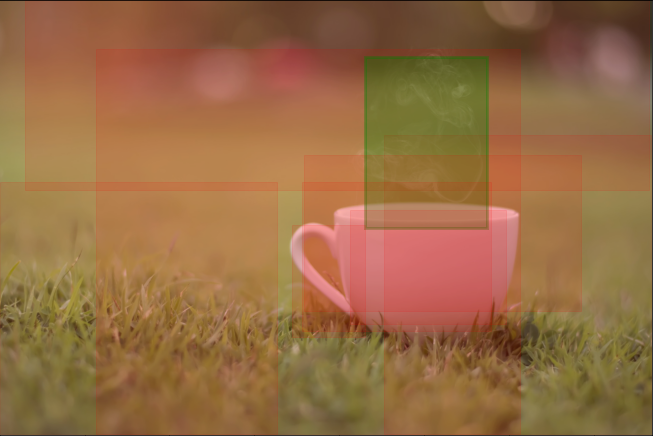} }}
    
    \caption{UNITER $S_{VL}$ attention for captions \textit{``a dog on a rock next to a person"} and \textit{``there is a mug in some grass"}. Shown are boxes that attend highly to the displayed token, with the maximally attending bounding box in green; others in red. Observe that although the argmax often does pick up on a relevant bounding box, it is prone to missing critical visual information, e.g. focusing on only the backpack in (b).}
    \label{fig:svl_ex}
\end{figure}

To illustrate how the soft approach accounts for critical cross-modal alignment information, we present a few Winoground examples with UNITER's cross-modal attention activations in Fig. \ref{fig:slv_ex} and \ref{fig:svl_ex}. We use UNITER since this is the baseline model from which attentional information is bootstrapped to calculate cross-modal alignments. For example, in Fig. \ref{fig:svl_ex}c, using the representation for the bounding box covering the mug's handle may not adequately capture the visual referent of `mug' and therefore disrupt our ability to calculate the visual-basis relation between `mug' and `grass' if restricted by an argmax.

\subsection{Quantitative}
In the absence of annotations, we attempted a quantitative measurement of whether overlap in argmaxes (several words attending to one bounding box or vice versa) as quantified by the Shannon Entropy of argmax indices inversely correlates with soft Winoground score. Intuitively, if an example has more like a one-to-one mapping between text and image, the entropy of its cross-modal argmaxes should be higher as each token will attend to a different box, which would suggest that the model is better aligning entities. However, we found no significant correlation with Winoground score, which we attribute to the fact that high entropy on its own doesn't mean \textit{correct} entity alignment. 

Rather, high entropy in argmax indices could still be produced by a bad representation if `mug' attends to the grass \& `grass' attends to the mug; conversely, low entropy could be produced by a good representation for an example like `fire truck' where two tokens refer to a single object. Quantitative exploration of cross-modal attention is difficult without annotations and we leave this task to future work to explore in a multimodal compositionality context.

As a general takeaway, while the cross-modal argmax assumption of IAIS does hold in some cases and may be more meaningful during the course of IAIS training, it is clearly quite a strict assumption that could suffer if an entity attends to several cross-modal entities or there are no corresponding cross-modal entities. Furthermore, since IAIS is only active in the final self-attention layer, all the token representations are intermixed and therefore don't necessarily have a one-to-one correspondence with our intuitive notions of what they should be—the word `turbine' may not solely represent the traditional meaning of that word but perhaps the entire scene that includes the turbines, clouds, and ground.

We hypothesize that by removing the hard argmax assumption, our approach better accounts for varying cross-modal entity equivalences and thus enables stronger relation alignment. By also calculating alignment between all pairs of source and target modality entities, CACR should considerably improve sample efficiency, which is important considering that the final layer $S$ matrix of the converged IAIS model is largely flat. Therefore it's important to backpropagate as much alignment knowledge over the course of training as possible, which CACR's soft equivalence weighting implicitly enables.

% Since another source of issues may be the lack of high quality visual tokens to align with, we pair SIAIS with VinVL, a recent visual feature extractor and the pre-IAIS SotA on Winoground.

\section{Conclusion}

In this work, we identified that a key factor holding back models from vision-language representational compositionality is cross-modal relation alignment. We categorized recent compositional inductive bias approaches into 3 categories: Structural Model, Structural Data, and Structural Training, showing that a previous Structural Training model (IAIS) achieves state-of-the-art performance on Winoground. We then identified a potential key weakness in IAIS, its hard argmax assumption, and developed a soft cross-modal equivalence approach to address it. Having linear algebraically simplified this approach, we arrived at CACR, an auxiliary loss that encourages cross-modal congruence of intra-modal attention. CACR improves on IAIS' performance on Winoground, and even outperforms a UNITER model nearly twice as large.

% An especially interesting future direction we'd like to explore is how attention flow \cite{abnar2020quantifying} captures compositional structure, and whether guiding attention 

As computational scaling becomes more widespread, it's necessary to develop compositional inductive biases that do not require complex annotated data or exotic model architectures. Our work illustrates how taking advantage of the transformer's own attentional structure can improve the quality of fine-grained vision-language representations, opening the avenue for large scale approaches to visually-grounded compositionality.

\section{Limitations, Risks \& Ethics}

Though CACR shows significant gains in compositional performance, results are limited in their exploration of only one pre-trained model and compositionality dataset. A significant risk of models is their tendency to be biased by distributions in their training data; vision-language models are not free from this flaw, but we see our work as teaching VLMs to learn better structured representations rather than memorizing spurious correlations in data. We remain far from solving the vision-language compositionality problem, so biases must continue to be actively mitigated.

\section*{Acknowledgements}

This material is based upon work partially supported by National Science Foundation awards 1722822 and 1750439, and National Institutes of Health awards R01MH125740, R01MH132225, R01MH096951 and R21MH130767. Any opinions, findings, conclusions, or recommendations expressed in this material are those of the author(s) and do not necessarily reflect the views of the sponsors, and no official endorsement should be inferred.

{\small
\bibliographystyle{plainnat}
\bibliography{custom,anthology}

\begin{thebibliography}{35}
\providecommand{\natexlab}[1]{#1}
\providecommand{\url}[1]{\texttt{#1}}
\expandafter\ifx\csname urlstyle\endcsname\relax
  \providecommand{\doi}[1]{doi: #1}\else
  \providecommand{\doi}{doi: \begingroup \urlstyle{rm}\Url}\fi

\bibitem[Abnar and Zuidema(2020)]{abnar2020quantifying}
Samira Abnar and Willem Zuidema.
\newblock Quantifying attention flow in transformers.
\newblock In \emph{Proceedings of the 58th Annual Meeting of the Association
  for Computational Linguistics}, pages 4190--4197, 2020.

\bibitem[Aflalo et~al.(2022)Aflalo, Du, Tseng, Liu, Wu, Duan, and
  Lal]{aflalo2022interpret}
Estelle Aflalo, Meng Du, Shao-Yen Tseng, Yongfei Liu, Chenfei Wu, Nan Duan, and
  Vasudev Lal.
\newblock Vl-interpret: An interactive visualization tool for interpreting
  vision-language transformers.
\newblock In \emph{Proceedings of the IEEE/CVF Conference on Computer Vision
  and Pattern Recognition}, pages 21406--21415, 2022.

\bibitem[Andreas et~al.(2016)Andreas, Rohrbach, Darrell, and
  Klein]{andreas2016neural}
Jacob Andreas, Marcus Rohrbach, Trevor Darrell, and Dan Klein.
\newblock Neural module networks.
\newblock In \emph{Proceedings of the IEEE conference on computer vision and
  pattern recognition}, pages 39--48, 2016.

\bibitem[Bai et~al.(2021)Bai, Wang, Chen, Yang, Bai, Yu, and
  Tong]{bai2021syntax}
Jiangang Bai, Yujing Wang, Yiren Chen, Yaming Yang, Jing Bai, Jing Yu, and
  Yunhai Tong.
\newblock Syntax-bert: Improving pre-trained transformers with syntax trees.
\newblock \emph{arXiv preprint arXiv:2103.04350}, 2021.

\bibitem[Bugliarello et~al.(2021)Bugliarello, Cotterell, Okazaki, and
  Elliott]{bugliarello-etal-2021-multimodal}
Emanuele Bugliarello, Ryan Cotterell, Naoaki Okazaki, and Desmond Elliott.
\newblock Multimodal pretraining unmasked: A meta-analysis and a unified
  framework of vision-and-language {BERT}s.
\newblock \emph{Transactions of the Association for Computational Linguistics},
  9:\penalty0 978--994, 2021.
\newblock \doi{10.1162/tacl_a_00408}.
\newblock URL \url{https://aclanthology.org/2021.tacl-1.58}.

\bibitem[Chen et~al.(2020)Chen, Li, Yu, El~Kholy, Ahmed, Gan, Cheng, and
  Liu]{chen2020uniter}
Yen-Chun Chen, Linjie Li, Licheng Yu, Ahmed El~Kholy, Faisal Ahmed, Zhe Gan,
  Yu~Cheng, and Jingjing Liu.
\newblock Uniter: Universal image-text representation learning.
\newblock In \emph{European conference on computer vision}, pages 104--120.
  Springer, 2020.

\bibitem[Cui et~al.(2021)Cui, Yu, Wang, Zhao, Zhang, Wang, and Yu]{cui21rosita}
Yuhao Cui, Zhou Yu, Chunqi Wang, Zhongzhou Zhao, Ji~Zhang, Meng Wang, and Jun
  Yu.
\newblock Rosita: Enhancing vision-and-language semantic alignments via
  cross-and intra-modal knowledge integration.
\newblock In \emph{Proceedings of the 29th ACM International Conference on
  Multimedia}, pages 797--806, 2021.

\bibitem[Guo et~al.(2019)Guo, Liu, Tang, Li, Luo, and Lu]{guo19vsua}
Longteng Guo, Jing Liu, Jinhui Tang, Jiangwei Li, Wei Luo, and Hanqing Lu.
\newblock Aligning linguistic words and visual semantic units for image
  captioning.
\newblock In \emph{Proceedings of the 27th ACM International Conference on
  Multimedia}, MM '19, page 765–773, New York, NY, USA, 2019. Association for
  Computing Machinery.
\newblock ISBN 9781450368896.
\newblock \doi{10.1145/3343031.3350943}.
\newblock URL \url{https://doi.org/10.1145/3343031.3350943}.

\bibitem[Hong et~al.(2021)Hong, Li, Zhu, and Huang]{hong2021vlgrammar}
Yining Hong, Qing Li, Song-Chun Zhu, and Siyuan Huang.
\newblock Vlgrammar: Grounded grammar induction of vision and language.
\newblock In \emph{Proceedings of the IEEE/CVF International Conference on
  Computer Vision (ICCV)}, pages 1665--1674, October 2021.

\bibitem[Ilinykh and Dobnik(2022)]{ilinykh-dobnik-2022-attention}
Nikolai Ilinykh and Simon Dobnik.
\newblock Attention as grounding: Exploring textual and cross-modal attention
  on entities and relations in language-and-vision transformer.
\newblock In \emph{Findings of the Association for Computational Linguistics:
  ACL 2022}, pages 4062--4073, Dublin, Ireland, May 2022. Association for
  Computational Linguistics.
\newblock \doi{10.18653/v1/2022.findings-acl.320}.
\newblock URL \url{https://aclanthology.org/2022.findings-acl.320}.

\bibitem[Khan et~al.(2022)Khan, BG, Yu, Schulter, Chandraker, and
  Fu]{khan2022simla}
Zaid Khan, Vijay~Kumar BG, Xiang Yu, Samuel Schulter, Manmohan Chandraker, and
  Yun Fu.
\newblock Single-stream multi-level alignment for vision-language pretraining.
\newblock \emph{arXiv preprint arXiv:2203.14395}, 2022.

\bibitem[Kim et~al.(2022)Kim, Song, and Zhang]{kim2022cross}
Taehyeong Kim, Hyeonseop Song, and Byoung-Tak Zhang.
\newblock Cross-modal alignment learning of vision-language conceptual systems.
\newblock \emph{arXiv preprint arXiv:2208.01744}, 2022.

\bibitem[Li et~al.(2020)Li, Zhou, Li, Xu, and Cao]{li2020improving}
Zhongli Li, Qingyu Zhou, Chao Li, Ke~Xu, and Yunbo Cao.
\newblock Improving bert with syntax-aware local attention.
\newblock \emph{arXiv preprint arXiv:2012.15150}, 2020.

\bibitem[Liang et~al.(2022)Liang, Zadeh, and Morency]{liang2022foundations}
Paul~Pu Liang, Amir Zadeh, and Louis-Philippe Morency.
\newblock Foundations and recent trends in multimodal machine learning:
  Principles, challenges, and open questions.
\newblock \emph{arXiv preprint arXiv:2209.03430}, 2022.

\bibitem[Loshchilov and Hutter(2017)]{loshchilov2017decoupled}
Ilya Loshchilov and Frank Hutter.
\newblock Decoupled weight decay regularization.
\newblock \emph{arXiv preprint arXiv:1711.05101}, 2017.

\bibitem[Mare{\v{c}}ek and Rosa(2019)]{marevcek2019balustrades}
David Mare{\v{c}}ek and Rudolf Rosa.
\newblock From balustrades to pierre vinken: Looking for syntax in transformer
  self-attentions.
\newblock \emph{arXiv preprint arXiv:1906.01958}, 2019.

\bibitem[Milewski et~al.(2022)Milewski, de~Lhoneux, and Moens]{milewski22}
Victor Milewski, Miryam de~Lhoneux, and Marie~Francine Moens.
\newblock Finding structural knowledge in multimodal-bert.
\newblock In \emph{Proceedings of the 60th Annual Meeting of the Association
  for Computational Linguistics (Volume 1: Long Papers)}, pages 5658--5671,
  2022.

\bibitem[Paszke et~al.(2019)Paszke, Gross, Massa, Lerer, Bradbury, Chanan,
  Killeen, Lin, Gimelshein, Antiga, et~al.]{paszke2019pytorch}
Adam Paszke, Sam Gross, Francisco Massa, Adam Lerer, James Bradbury, Gregory
  Chanan, Trevor Killeen, Zeming Lin, Natalia Gimelshein, Luca Antiga, et~al.
\newblock Pytorch: An imperative style, high-performance deep learning library.
\newblock \emph{Advances in neural information processing systems}, 32, 2019.

\bibitem[Radford et~al.(2021)Radford, Kim, Hallacy, Ramesh, Goh, Agarwal,
  Sastry, Askell, Mishkin, Clark, et~al.]{radford2021clip}
Alec Radford, Jong~Wook Kim, Chris Hallacy, Aditya Ramesh, Gabriel Goh,
  Sandhini Agarwal, Girish Sastry, Amanda Askell, Pamela Mishkin, Jack Clark,
  et~al.
\newblock Learning transferable visual models from natural language
  supervision.
\newblock In \emph{International Conference on Machine Learning}, pages
  8748--8763. PMLR, 2021.

\bibitem[Ren et~al.(2021)Ren, Lin, Zhao, Men, Yang, Zhou, Sun, and
  Yang]{ren21iais}
Shuhuai Ren, Junyang Lin, Guangxiang Zhao, Rui Men, An~Yang, Jingren Zhou,
  Xu~Sun, and Hongxia Yang.
\newblock Learning relation alignment for calibrated cross-modal retrieval.
\newblock In \emph{Proceedings of the 59th Annual Meeting of the Association
  for Computational Linguistics and the 11th International Joint Conference on
  Natural Language Processing (Volume 1: Long Papers)}, pages 514--524, 2021.

\bibitem[Tan and Bansal(2019)]{tan2019lxmert}
Hao Tan and Mohit Bansal.
\newblock Lxmert: Learning cross-modality encoder representations from
  transformers.
\newblock In \emph{Proceedings of the 2019 Conference on Empirical Methods in
  Natural Language Processing and the 9th International Joint Conference on
  Natural Language Processing (EMNLP-IJCNLP)}, pages 5100--5111, 2019.

\bibitem[Thrush et~al.(2022)Thrush, Jiang, Bartolo, Singh, Williams, Kiela, and
  Ross]{thrush22}
Tristan Thrush, Ryan Jiang, Max Bartolo, Amanpreet Singh, Adina Williams, Douwe
  Kiela, and Candace Ross.
\newblock Winoground: Probing vision and language models for visio-linguistic
  compositionality.
\newblock In \emph{Proceedings of the IEEE/CVF Conference on Computer Vision
  and Pattern Recognition}, pages 5238--5248, 2022.

\bibitem[Wan et~al.(2021)Wan, Han, Zheng, and Tuytelaars]{wan2021cliora}
Bo~Wan, Wenjuan Han, Zilong Zheng, and Tinne Tuytelaars.
\newblock Unsupervised vision-language grammar induction with shared structure
  modeling.
\newblock In \emph{International Conference on Learning Representations}, 2021.

\bibitem[Wang et~al.(2022{\natexlab{a}})Wang, Yasunaga, Ren, Wada, and
  Leskovec]{wang22vqagnn}
Yanan Wang, Michihiro Yasunaga, Hongyu Ren, Shinya Wada, and Jure Leskovec.
\newblock Vqa-gnn: Reasoning with multimodal semantic graph for visual question
  answering.
\newblock \emph{arXiv preprint arXiv:2205.11501}, 2022{\natexlab{a}}.

\bibitem[Wang et~al.(2022{\natexlab{b}})Wang, You, Li, Zareian, Park, Liang,
  Chang, and Chang]{wang2022sgeitl}
Zhecan Wang, Haoxuan You, Liunian~Harold Li, Alireza Zareian, Suji Park, Yiqing
  Liang, Kai-Wei Chang, and Shih-Fu Chang.
\newblock Sgeitl: Scene graph enhanced image-text learning for visual
  commonsense reasoning.
\newblock In \emph{Proceedings of the AAAI Conference on Artificial
  Intelligence}, volume~36, pages 5914--5922, 2022{\natexlab{b}}.

\bibitem[Wu et~al.(2019)Wu, Mao, Zhang, Jiang, Li, Sun, and Ma]{Wu_2019_univse}
Hao Wu, Jiayuan Mao, Yufeng Zhang, Yuning Jiang, Lei Li, Weiwei Sun, and
  Wei-Ying Ma.
\newblock Unified visual-semantic embeddings: Bridging vision and language with
  structured meaning representations.
\newblock In \emph{Proceedings of the IEEE/CVF Conference on Computer Vision
  and Pattern Recognition (CVPR)}, June 2019.

\bibitem[Xue et~al.(2021)Xue, Huang, Liu, Peng, Fu, Li, and Luo]{xue2021imf}
Hongwei Xue, Yupan Huang, Bei Liu, Houwen Peng, Jianlong Fu, Houqiang Li, and
  Jiebo Luo.
\newblock Probing inter-modality: Visual parsing with self-attention for
  vision-and-language pre-training.
\newblock \emph{Advances in Neural Information Processing Systems},
  34:\penalty0 4514--4528, 2021.

\bibitem[Yang et~al.(2021{\natexlab{a}})Yang, Gao, Zhang, and Cai]{yang21apn}
Xu~Yang, Chongyang Gao, Hanwang Zhang, and Jianfei Cai.
\newblock Auto-parsing network for image captioning and visual question
  answering.
\newblock In \emph{Proceedings of the IEEE/CVF International Conference on
  Computer Vision}, pages 2197--2207, 2021{\natexlab{a}}.

\bibitem[Yang et~al.(2021{\natexlab{b}})Yang, Zhang, Qi, and Cai]{yang21catt}
Xu~Yang, Hanwang Zhang, Guojun Qi, and Jianfei Cai.
\newblock Causal attention for vision-language tasks.
\newblock In \emph{Proceedings of the IEEE/CVF Conference on Computer Vision
  and Pattern Recognition (CVPR)}, pages 9847--9857, June 2021{\natexlab{b}}.

\bibitem[Young et~al.(2014)Young, Lai, Hodosh, and
  Hockenmaier]{young2014flickr}
Peter Young, Alice Lai, Micah Hodosh, and Julia Hockenmaier.
\newblock From image descriptions to visual denotations: New similarity metrics
  for semantic inference over event descriptions.
\newblock \emph{Transactions of the Association for Computational Linguistics},
  2:\penalty0 67--78, 2014.

\bibitem[Yu et~al.(2021)Yu, Tang, Yin, Sun, Tian, Wu, and Wang]{yu20ernievil}
Fei Yu, Jiji Tang, Weichong Yin, Yu~Sun, Hao Tian, Hua Wu, and Haifeng Wang.
\newblock Ernie-vil: Knowledge enhanced vision-language representations through
  scene graphs.
\newblock In \emph{Proceedings of the AAAI Conference on Artificial
  Intelligence}, volume~35, pages 3208--3216, 2021.

\bibitem[Yuksekgonul et~al.(2022)Yuksekgonul, Bianchi, Kalluri, Jurafsky, and
  Zou]{yuksekgonul2022bags}
Mert Yuksekgonul, Federico Bianchi, Pratyusha Kalluri, Dan Jurafsky, and James
  Zou.
\newblock When and why vision-language models behave like bag-of-words models,
  and what to do about it?
\newblock \emph{arXiv preprint arXiv:2210.01936}, 2022.

\bibitem[Zhang(2022)]{zhang-2022-improve}
Bryan Zhang.
\newblock Improve {MT} for search with selected translation memory using search
  signals.
\newblock In \emph{Proceedings of the 15th Biennial Conference of the
  Association for Machine Translation in the Americas (Volume 2: Users and
  Providers Track and Government Track)}, pages 123--131, Orlando, USA,
  September 2022. Association for Machine Translation in the Americas.
\newblock URL \url{https://aclanthology.org/2022.amta-upg.9}.

\bibitem[Zhang and Peng(2019)]{zhang2019agha}
Junchao Zhang and Yuxin Peng.
\newblock Hierarchical vision-language alignment for video captioning.
\newblock In \emph{International Conference on Multimedia Modeling}, pages
  42--54. Springer, 2019.

\bibitem[Zhang et~al.(2021)Zhang, Li, Hu, Yang, Zhang, Wang, Choi, and
  Gao]{zhang2021vinvl}
Pengchuan Zhang, Xiujun Li, Xiaowei Hu, Jianwei Yang, Lei Zhang, Lijuan Wang,
  Yejin Choi, and Jianfeng Gao.
\newblock Vinvl: Revisiting visual representations in vision-language models.
\newblock In \emph{Proceedings of the IEEE/CVF Conference on Computer Vision
  and Pattern Recognition}, pages 5579--5588, 2021.

\end{thebibliography}
}

\end{document}